\documentclass{llncs}

\usepackage{epsfig}
\usepackage{amsfonts}
\usepackage{amssymb}
\usepackage{amsmath}
\usepackage{url}
\usepackage{algorithmic,algorithm}

\newcommand{\lpath}[0]{$k$-lp$^2$}
\newcommand{\mpath}[0]{$k$-mp$^2$}

\begin{document}

\mainmatter

\title{On the Effect of Connectedness for\\Biobjective Multiple and Long Path Problems}
\titlerunning{Connectedness for Biobjective Multiple and Long Path Problems}

\author{S\'ebastien Verel\inst{1,2} \and Arnaud Liefooghe\inst{1,3} \and J\'er\'emie Humeau\inst{1,4} \and\\Laetitia Jourdan\inst{1} \and Clarisse~Dhaenens\inst{1,3}}

\authorrunning{S. Verel, A. Liefooghe, J. Humeau, L. Jourdan, C. Dhaenens}   

\institute{INRIA Lille-Nord Europe, France 
\and Universit\'e Nice Sophia Antipolis, I3S -- CNRS, France
\and Universit\'e Lille 1, LIFL -- CNRS, France\\
\and \'Ecole des Mines de Douai, IA department, France\\
\email{verel@i3s.unice.fr, arnaud.liefooghe@univ-lille1.fr, jeremie.humeau@mines-douai.fr, laetitia.jourdan@inria.fr, clarisse.dhaenens@lifl.fr}
}

\maketitle

\begin{abstract}
Recently, the property of connectedness has been claimed to give a strong motivation on the design of local search techniques
for multiobjective combinatorial optimization.
Indeed, when connectedness holds, a basic Pareto local search, initialized with at least one non-dominated solution,
allows to identify the efficient set exhaustively.
However,
this becomes quickly infeasible in practice as the number of efficient solutions typically grows exponentially with the instance size.
As a consequence, we generally have to deal with a limited-size approximation,
ideally a representative sample of efficient solutions.
In this paper, we propose the biobjective long and multiple path problems. 
We show experimentally that, on the first problem, even if the efficient set is connected,
a local search may be outperformed by a simple evolutionary algorithm in the sampling of the efficient set.
At the opposite, on the second problem, a local search algorithm may successfully approximate a disconnected efficient set.
Then, we argue that connectedness is not the single property to study for the design of multiobjective local search algorithms.
This work opens new discussions on a proper definition of multiobjective fitness landscapes.
\end{abstract}

\section{Introduction}
\label{sec:intro}

The single-objective long path problem \cite{horn1994} has been introduced to show that a problem instance can be difficult to solve for a hillclimber-like heuristic
even if the search space is unimodal, \textit{i.e.} the single local optimum is the global optimum.
For such a problem, a hillclimber guarantees to reach the global optimum,
but the length of the path to get it is exponential in the dimension of the search space.
As a consequence, a hillclimbing-based heuristic cannot expect to solve the problem in polynomial time.
The `path length' takes then place in the rank of problem difficulty, on the same level as multimodality, ruggedness, deceptivity, and so on.
Rudolph \cite{Rudolph96} demonstrated
that the long path problem can be solved in a polynomial expected amount of time for a $(1+1)$~evolutionary algorithm~(EA) which is able to mutate more than one bit at a time.
This $(1+1)$ EA is able to take some shortcuts on the outside of the path so that it makes the computation more efficient.
However, it does not change the argument that, even for unimodal problems, 
the path length to the global optimum must be taken into account in the design of efficient local search algorithms.

Like in single-objective optimization, the structure of the search space can explain the difficulty for multiobjective local search methods.
In multiobjective combinatorial optimization (MoCO), the efficient set is the set of solutions which are not dominated by any other feasible solution.
It is often claimed that the structure of this efficient set plays a crucial role for the development of efficient local search methods \cite{gorski2006b}.
Connectedness is related to the property that efficient solutions are connected (at distance $1$)
with respect to a neighborhood relation \cite{ehrgott1997}.
This property has later been extended to the notion of cluster, where distances can take higher values \cite{paquete2009}.
When connectedness holds, it becomes possible to find all the efficient solutions by means of the iterative exploration of the neighborhood of the current approximation set
by starting by one (or more) solution(s) from the efficient set.
This strategy coincides with the Pareto Local Search (PLS) algorithm \cite{paquete2004}, initialized with one efficient solution,
and then acts like an exact approach.
However, a common knowledge is that, for most MoCO problems, the number of non-dominated solutions is not polynomial in the size of the problem instance~\cite{ehrgott2005},
so that a PLS algorithm can take an exponential time to identify the efficient set once the later contains an exponential number of solutions. 
Then, the goal of the optimization process is often to identify a representative sample set, containing a limited number of efficient solutions.

In this work, we argue that connectedness is not the only feature which explains the difficulty of MoCO for search algorithms.
Analogously to the single-objective long path problems,
where a hillclimbing algorithm is outperformed by a simple EA, even if the search space is unimodal,
we here oppose straightforward extensions of those algorithms, a hillclimbing algorithm and a simple EA, in a multiobjective context.
On one side, PLS extends a single-objective hillclimber in terms of Pareto dominance \cite{paquete2004}.
At the opposite, we use an adaptation of the Simple Evolutionary Multiobjective Optimization (SEMO) algorithm~\cite{Laumanns04}.
Both approaches are initialized with one solution from the efficient set, corresponding to an extreme point of the Pareto front. 
In this paper, we propose the definition of the biobjective long path problem (\lpath) and of the biobjective multiple path problem (\mpath).
With \lpath, we show experimentally that, even if the efficient set is connected,
the runtime required by PLS to find a reasonably good approximation (in terms of hypervolume~\cite{zitzler1999}) is larger than for SEMO,
and becomes computationally prohibitive for large-size instances.
Furthermore, we construct \mpath~instances where the efficient set is completely disconnected,
but some additional shortcuts are available to walk from one non-dominated solution to the others.
In this case, we show experimentally that PLS can find a good approximation in a significantly less amount of time than SEMO.
Indeed, both algorithms differ in the way they sample the efficient set.
For \lpath, PLS can only follow the path defined by the connectedness property while SEMO is able to take some shortcuts outside of the path.
For \mpath, PLS takes advantage of the multiple paths, defined outside the efficient set, which are temporally non-dominated and that lead to further non-dominated solutions.

The reminder of the paper is organized as follows.
First, some notions related to MoCO, connectedness and long path problems are briefly presented in the next section.
Section~\ref{sec:longpath} introduces the class of biobjective long path problems, for which the efficient set is fully connected and exponential in the size of the problem instance.
Next, the class of multiple path problems is presented in Section~\ref{sec:multiple}.
It handles an exponential number of disconnected efficient solutions.
Our experiments illustrate that PLS appears to be outperformed by SEMO for biobjective long path problems, 
while more surprisingly, the opposite occurs for multiple path problems.
This work leads to further investigations on a proper definition of fitness landscapes for MoCO,
not only with regards to the efficient set itself, but also to the way that leads to its approximation.

\section{Background}

\subsection{Multiobjective Combinatorial Optimization}

A multiobjective optimization problem can be defined by a set of $m \geq 2$ objective functions $(f_1, f_2,\dots, f_m)$,
and a set $X$ of feasible solutions in the \emph{decision space}.
In the combinatorial case, $X$ is a discrete set.
Let $Z=f(X)$ denote the set of feasible outcome vectors in the \emph{objective space}.
To each solution $x \in X$ is assigned an objective vector on the basis of a vector function 
$f : X \rightarrow Z$ with $f(x) = ( f_1(x) , f_2(x) , \ldots , f_m(x) )$.
Without loss of generality, we here assume that all $m$ objective functions are to be maximized.  
A solution $x \in X$ is said to \emph{dominate} a solution $x' \in X$, denoted by $x \succ x'$,
iff $\forall i \in \{1,2,\dots,m\}$, $f_i(x) \geq f_i(x')$ and $\exists j \in \{1,2,\dots,m\} $ such as $f_j(x) > f_j(x')$.
A solution $x \in X$ is said to be \emph{efficient} (or \emph{Pareto optimal}, \emph{non-dominated})
if there does not exist any other solution $x^{'} \in X$ such that $x^{'}$ dominates $x$.
The set of all efficient solutions is called the \emph{efficient set}
and its mapping in the objective space is called the \emph{Pareto front}.
A possible approach in MoCO is to find a minimal set of efficient solutions, such that strictly one solution maps to each non-dominated vector.
However, generating the entire efficient set of a MoCO problem is usually infeasible for two main reasons.
First, the number of efficient solutions is typically exponential in the size of the problem instance \cite{ehrgott2005}.
In that sense, most MoCO problems are said to be intractable.
Second, deciding if a feasible solution belongs to the efficient set is known to be NP-complete for numerous MoCO problems~\cite{serafini1986},
even if none of its single-objective counterpart is NP-hard.
Therefore, the overall goal is often to identify a good efficient set approximation, ideally a subpart of the efficient~set.
To this end, heuristic approaches have received a growing interest in the last decades.

\subsection{Local Search and Connectedness}
\label{sec:ls}
A \emph{neighborhood structure} is a function $\mathcal{N} : X \rightarrow 2^X$ that assigns a set of solutions $\mathcal{N}(x) \subset X$ to any solution $x \in X$.
$\mathcal{N}(x)$ is called the \emph{neighborhood} of $x$, and a solution $x' \in \mathcal{N}(x)$ is called a \emph{neighbor} of $x$.
Local search algorithms for MoCO, like the Pareto Local Search (PLS) \cite{paquete2004}, generally combine the use of such a neighborhood structure
with the management of an archive (or population) of mutually non-dominated solutions found so far.
The basic idea is to iteratively improve this archive by exploring the neighborhood of its own content until no further improvement is possible,
or until another stopping condition is fulfilled.

Recently, local search approaches have been successfully applied to MoCO problems.
Some structural properties of the landscape seem to allow the search space to be explored in an effective way.
Such a property, related to the efficient set, is \emph{connectedness}~\cite{gorski2006b,ehrgott1997}.
As argued by the original authors, 
it could provide a theoretical justification for the design of multiobjective local search.
Let us define a graph such that each node represents an efficient solution,
and an edge connects a pair of nodes if the corresponding solutions are neighbors with respect to a given neighborhood relation \cite{ehrgott1997}.
The efficient set is said to be \emph{connected} if there exists a path between every pair of nodes in the graph.
Paquete and St\"utzle \cite{paquete2009} extended this notion
by introducing an arbitrary distance separating two efficient solutions
({\itshape i.e.} the minimal number of neighbors to visit to go from one solution to another).
Unfortunately, in the general case, rather negative results have been reported in the literature for some classical MoCO problems
\cite{gorski2006b,ehrgott1997}.
However, in practice, many empirical results show that efficient solutions for some MoCO problems are strongly clustered
with respect to more classical neighborhood structures from combinatorial optimization, see for instance~\cite{paquete2009}.
Indeed, in the case of connectedness, by starting with one or more non-dominated solutions,
it becomes possible to find all the efficient solutions through a basic iterative neighborhood exploration procedure, like PLS.
However, we show in this paper that connectedness is not the only property to deal with when searching for an
approximation of the efficient set.

\subsection{The Single-objective Long $k$-path Problem}

The long path problem has been introduced by Horn et al. \cite{horn1994} to design unimodal landscapes
where the path length to reach the global optimum is exponential in the size of the problem instance.
The long $k$-path is defined on bit strings of size~$l$.
Let $P_{l,k}$ be a long $k$-path of dimension $l$, and $P_{l,k}(i)$ the $i^{th}$ solution on this path.
The long $k$-path of dimension $1$ is only made of two solutions $P_{1,k} = (0, 1)$, and the path of dimension $l+k$ can be defined by recursion:
$$
P_{l+k,k}(i) =
\left\lbrace
\begin{array}{l l}
0^k P_{l,k}(i)  & \text{if } 0 \leq i < s_{l,k} \\
0^{k-j}1^j P_{l,k}(s_{l,k} - 1)  & \text{if } s_{l,k} \leq i < s_{l,k} + k - 1 \text{ with } j = i - s_{l,k} + 1 \\
1^k P_{l,k}(s_{l+k,k} - 1 - i) & \text{if } s_{l,k} + k - 1 \leq i < s_{l+k,k}\\
\end{array}
\right.
$$
where $s_{l,k} = |P_{l,k}| = 2 s_{l-k,k} + (k-1) = (k+1) 2^{(l-1)/k} - k + 1$ is the length of the $k$-path of dimension $l$.
The fitness function of the long $k$-path problem (to be maximized) is defined as follows.
For all $x \in \{0,1\}^l$:
$$
f(x) =
\left\lbrace
\begin{array}{l l}
l + i & \text{if } x \in P_{l,k} \text{ and } x=P_{l,k}(i) \\
|x|_{0}  & \text{if } x \not\in P_{l,k} \\
\end{array}
\right.
$$
where $|x|_{0}$ is the number of `$0$' in the bit string $x$.
In the long $k$-path, a shortcut can be found by flipping $k$ consecutive bits.
For a hillclimbing algorithm which chooses the best solution in the neighborhood defined by Hamming distance~$1$, 
the number of iterations to reach the global optimum matches the length of the path, $s_{l,k}$. 
The number of evaluations is then ($l \cdot s_{l,k}$) for a hillclimber. 
On the contrary, a $(1+1)$ EA which flips each bit with a probability $p=1/l$ at each iteration
is found the global optimum in polynomial expected running time ${\cal O}(l^{k+1}/k)$~\cite{Rudolph96}%
\footnote{The lower bound of the expected runtime could be exponential when $k=\sqrt{l-1}$~\cite{droste1998}.}.

\section{The Biobjective Long $k$-path Problem}
\label{sec:longpath}

In this section, we propose a biobjective problem where the efficient set is connected,
but so huge that the full enumeration of it cannot be made in polynomial time.
We define the \emph{biobjective long $k$-path problem} to show that the required runtime to sample a connected efficient set
can be very long for a simple local search algorithm.

\subsection{Definition}

The biobjective long $k$-path problem (\lpath) is defined on a bit string of length~$l$, with an objective function vector of dimension $2$.
Each objective function corresponds to a `single' long $k$-path problem, which is to be maximized.
The \lpath~is built such that the efficient set matches the path $P_{l,k}$.
The objective function vector of \lpath~is defined as follows.
For all $x \in \{0,1\}^l$:
$$
f(x) = (f_1(x), f_2(x)) =
\left\lbrace
\begin{array}{l l}
h_{l,k}(i) & \text{if } x \in P_{l,k} \text{ and } x=P_{l,k}(i) \\
(|x|_{0}, |x|_{0})  & \text{if } x \not\in P_{l,k} \\
\end{array}
\right.
$$
where $h$ is the function which associates each integer $i$ to the point of coordinates~$(l + i, l + s_{l,k} - 1 - i)$ in the objective space.
So, the first objective is the fitness function of the single-objective long $k$-path problem.

\begin{figure} 
\begin{center}
\includegraphics[width=0.63\textwidth]{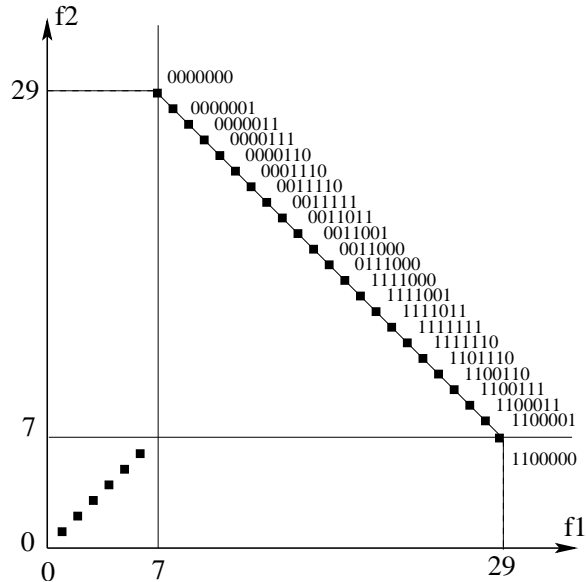} 
\caption{Objective space of the biobjective long $2$-path problem of dimension $l=7$. \label{fig:repr_path}}
\end{center}
\end{figure}

The efficient set of \lpath~corresponds to the path $P_{l,k}$ (see Fig. \ref{fig:repr_path}). 
By construction, all solutions in $P_{l,k}$ are neighbors with respect to Hamming distance $1$, so that the efficient set is connected.
The size of $P_{l,k}$ is $s_{l,k}=(k+1) 2^{(l-1)/k} - k + 1$, which cannot be enumerated in a polynomial number of evaluations
in the general case.
The efficient set of \lpath~is then ($i$) connected and ($ii$)~intractable.
Let us now experimentally examine the ability of search algorithms to identify a good approximation of it.

\subsection{Experimental Analysis}

\subsubsection{Ingredients.}
For the single-objective long path problems, existing studies are based on the comparison of a hillclimber and of a $(1+1)$ EA \cite{Rudolph96}.
Then, we will here consider straightforward multiobjective extensions of these approaches, respectively a PLS- and a SEMO-like algorithm.
They are both adapted to the path problems (\lpath~and \mpath) introduced in this paper,
and they will be respectively denoted by PLS$_p$ and SEMO$_p$ to differentiate them from their original implementation.
A pseudo-code is given in Algorithm \ref{algo:pls} and Algorithm \ref{algo:semo}, respectively.
At each PLS$_p$ iteration, one solution is chosen at random from the archive.
All solutions located at Hamming distance $1$ are evaluated and are checked for insertion in the archive.
For the problem under study, note that at most two neighbors are located on the long path,
with one of them being already found at a previous iteration.
The current solution is then marked as \emph{visited} in order to avoid a useless revaluation of its neighborhood.
At each SEMO$_p$ step, one solution is randomly chosen from the archive.
Each bit of this solution is independently flipped with a probability $p=1/l$, and the obtained solution is checked for insertion in the archive.
In PLS$_p$, the whole neighborhood is explored while in SEMO$_p$, all solutions are potentially reachable with respect to different probabilities%
\footnote{In SEMO, the neighborhood operator is generally supposed to be ergodic~\cite{Laumanns04}.}.
In order to take advantage of the connectedness property,
the archive of both algorithms is initialized with one solution from the efficient set: the bit string~$(0, 0, \dots, 0)$ of size $l$.

\begin{algorithm}
\caption{PLS$_p$}
\label{algo:pls}
\begin{algorithmic}
\STATE $A \leftarrow \{ 0^l \}$
\REPEAT
    \STATE select $x \in A$ at random such that $x$ is not \textit{visited}
    \STATE set $x$ to \textit{visited}
    \FORALL{$x^\prime$ such that $|x - x^\prime|_1=1$}
       \STATE updateArchive ($A$, $x^\prime$)
    \ENDFOR
\UNTIL{$I_H^\star - I_H(A) < \epsilon \cdot I_H^\star$}
\end{algorithmic}
\end{algorithm}

\begin{algorithm}
\caption{SEMO$_p$}
\label{algo:semo}
\begin{algorithmic}
\STATE $A \leftarrow \{ 0^l \}$
\REPEAT
    \STATE select $x \in A$ at random
    \STATE create $x^\prime$ by flipping each bit of $x$ with a probability $p=1/l$
    \STATE updateArchive ($A$, $x^\prime$)
\UNTIL{$I_H^\star - I_H(A) < \epsilon \cdot I_H^\star$}
\end{algorithmic}
\end{algorithm}

However, the efficient set of \lpath~is intractable.
It becomes then impracticable to use an unbounded archive for large-size problem instances.
As a consequence, contrary to the original approaches, we here maintain a \emph{bounded archive} of size $M$ in our implementation of the algorithms.
Our attempt is not to compare different bounded archiving techniques, 
but rather to limit the number of evaluations required for computing a reasonably good approximation of the efficient set.
So, we define a nearly ideal archiving method to find such an approximation for the particular case of \lpath.
If the Pareto front was linear, 
an `optimal' approximation of size $M$ contains uniformly distributed points over the segment~$[ (l,l+s_{l,k}-1) , (l+s_{l,k}-1,l) ]$
in the objective space.
Note that, in our case, those points do not necessarily correspond to feasible solutions in the decision space.
The distance between $2$ solutions with respect to the first objective is then $\delta = (s_{l,k} - 1) / (M-1)$.
The bounded archiving technique under consideration is given in Algorithm \ref{algoArchive}.
First, dominated solutions are always discarded.
If the number of non-dominated solutions becomes too large, the solution with the lowest first objective value
 which is too close from the previous one ({\itshape i.e.} the difference with respect to the first objective is below $\delta$)
is removed from the archive.
If this rule does not hold for any solution, the penultimate solution (with respect to the order defined by objective~$1$)
is removed (not the last one).
Of course, such an archiving technique is \lpath-specific,
but it does not introduce any bias within heuristic rules generally defined by existing diversity-based archiving approaches.

\begin{algorithm}
\begin{multicols*}{2}
\begin{algorithmic}
\STATE \textbf{updateArchive}($A$, $x$):
\FORALL{$a \in A$}
    \IF{$x \succ a$}
            \STATE $A \leftarrow A \setminus \{ a \}$
    \ENDIF
\ENDFOR
\IF{not $\exists a \in A : a \succ x$}
     \STATE $A \leftarrow A \cup \{ x \}$
   \IF{$|A| > M$}
      \STATE reduceArchive($A$)
   \ENDIF
\ENDIF
\end{algorithmic}
\columnbreak
\begin{algorithmic}
\STATE \textbf{reduceArchive}($A$):
\STATE Sort $A$ in the increasing order w.r.t $f_1$-values: $A = \{ a_1, a_2, a_3, \ldots \}$
\STATE $i \leftarrow 2$
\WHILE{$|A| > M$}
	\IF {$i = |A|$}
		\STATE $A \leftarrow A \setminus \{ a_{|A|-1} \}$
	\ELSIF{$f_1(a_{i}) - f_1(a_{i-1}) < \delta$}
		\STATE $A \leftarrow A \setminus \{ a_i \}$
	\ELSE
    		\STATE $i \leftarrow i + 1$
     \ENDIF
\ENDWHILE
\end{algorithmic}
\end{multicols*}
\caption{Bounded archiving}
\label{algoArchive}
\end{algorithm}

\subsubsection{Experimental Design.}
The algorithms are compared in terms of the required number of evaluations to attain a reasonable approximation of the efficient set.
The cost related to archiving is then ignored, as we want to focus on the complexity of algorithms independently of the archiving strategy.
The stopping criteria is based on a percentage of hypervolume $I_H$~\cite{zitzler1999} covered by the solutions from the archive.
For \lpath, an upper bound of the maximal hypervolume ($I^\star_H$) for an approximation of size $M$ can be computed
by uniformly distributing $M$ points over the Pareto front,
that is $I_H^\star = \delta^2 (M+1) M / 2$, $(l,l)$~being the reference point.
Once the hypervolume covered by the current archive $I_H(A)$ is below an $\epsilon$-value from $I_H^\star$, the algorithm stops.

The experimental study has been conducted with $k=2$ and dimensions $l = \{ 19, 29, 39, 49, 59 \}$.
We use an archive of size $M=100$,
and the required approximation to be found is less than $\epsilon = 2 \%$ of the maximal hypervolume.
In other words, at least $98\%$ of the best-possible approximation is covered in terms of hypervolume.
The archive is initialized with a bit string where all bits are set to `$0$'.
The number of evaluations is reported over $30$ independent runs.

\subsubsection{Results and Discussion.}
\begin{figure} 
\begin{center}
\includegraphics[width=0.7\textwidth]{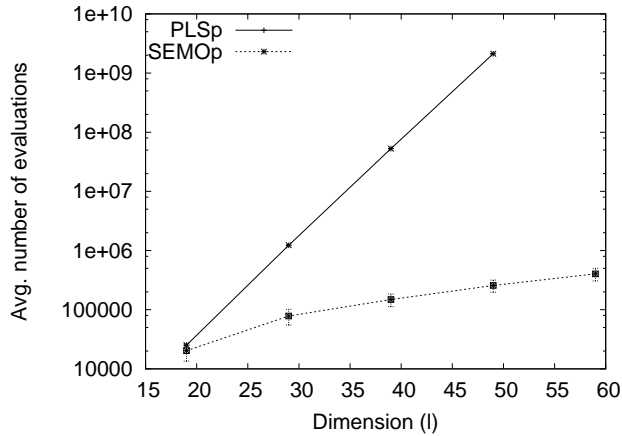}
\caption{Average value and standard deviation of the number of evaluations for PLS$_p$ and SEMO$_p$ 
on biobjective long $2$-path problems (log y-scale). \label{fig:longpath}}
\end{center}
\end{figure}

Fig. \ref{fig:longpath} shows the average and the standard deviation of the number of evaluations for each algorithm.
The number of evaluations required by PLS$_p$ seems to grow exponentially with the dimension $l$.
It could be interpreted as follows.
To approximate the efficient set, PLS$_p$ follows the long path.
When the archive reaches its maximum size,
the archiving technique let one solution at an `optimal' position in the objective space at every $\delta$ iteration.
So, at a given iteration $i$, the current hypervolume is approximately
$I_H(A) \approx \delta^2 (2M + 1 - j) \cdot j / 2$, where $j = \lceil i / \delta \rceil$.
Then, the stopping criteria is reached at the end of the long path only,
so that the number of evaluations is more than exponential in the dimension of the problem instance ($l$ times larger).
For SEMO$_p$,
the number of evaluations increases from $20.10^3$ evaluations for $l=19$ to $250.10^3$ for $l=59$.
The computational effort required by SEMO$_p$ and by PLS$_p$ is different of several orders of magnitude.
For SEMO$_p$, it is difficult to pretend that the runtime is polynomial or not,
nevertheless the number of evaluations remains huge.
The increase is higher than quadratic and seems to fit a cubic~curve.

To summarize, SEMO$_p$ can sample the efficient set more easily than PLS$_p$ by taking shortcuts out of the long path.
From the SEMO$_p$ point of view,
the efficient set is $k$-connected~\cite{paquete2009}:
one efficient solution can be reached by flipping $k$ bits of another efficient solution.
The computational difference between the two algorithms can be explained by different structures of the graph of efficient solutions.
For PLS$_p$, it is linear,
and for SEMO$_p$, the distance between $2$ efficient solutions in the graph is much smaller than the distance in the objective space.
This result suggests that the connectedness property is not fully satisfactorily to explain the degree of difficulty of the problem.
The structure of the graph of efficient solutions induced by the neighborhood relation should also be taken into account.
In the next section, we will show that the structure of this graph is still not enough to explain all the difficulties.

\section{The Biobjective Multiple $k$-path Problem}
\label{sec:multiple}

In the biobjective long $k$-path, the efficient set is connected, intractable and difficult to sample.
In this section, we define the biobjective multiple $k$-path problem~(\mpath) where the efficient set is still intractable but not connected anymore, 
while easier to sample for a PLS-like algorithm.

\subsection{Definition}

The idea is to modify \lpath~in order to make the efficient set disconnected (with respect to Hamming distance $1$),
and to add some shortcuts out of the path that guide the search towards efficient solutions.
A \mpath~instance of dimension~$l$ is defined for bit strings of size $l$ such that $(l-1)/k \in \mathbb{N}$, with $k$ being an even integer value.
First, let us define the additional paths, called \textit{extra paths}.
Let $D_{l,k}$ and $U_{l,k}$ be the extra paths of the $k$-path of dimension $l$. 
Let $u \in ( 0^k | 1^k )^{*}$ be a concatenation of $1^k$ and $0^k$.
$D_{l,k}(u,j,i)$ (resp. $U_{l,k}(u,j,i)$) is the $j^{th}$ solution on the extra path from solution $P_{l,k}(i_0) = u 0^k P_{l-|u|-k,k}(i)$ to solution $P_{l,k}(i_1) = u 1^k P_{l-|u|-k,k}(i)$ of the long $k$-path (resp. from $P_{l,k}(i_1)$ to $P_{l,k}(i_0)$). 
$D$ and $U$ are defined like the bridges in the single-objective long path problem~\cite{horn1994}.
$\forall p \in [ 0 .. \frac{l-1-k}{k} ] \ , \ \forall u \in ( 0^k | 1^k )^{p} \ , \ \forall i \in [0..s_{l-(p+1)k,k}-1] \ , \ \forall j \in [1..k-1]$:
$$
\left\lbrace 
\begin{array}{lcl}
D_{l,k}(u,j,i) & = & u 0^{k-j} 1^{j} P_{l-(p+1)k,k}(i) \\
U_{l,k}(u,j,i) & = & u 1^{k-j} 0^{j} P_{l-(p+1)k,k}(i) \\ 
\end{array}
\right.
$$
The sequence of neighboring solutions $( D_{l,k}(u,1,i), \ldots , D_{l,k}(u,k-1,i) )$ is the extra path to go from solution $P_{l,k}(i_{0})$ to solution $P_{l,k}(i_{1})$.
Respectively, the sequence $( U_{l,k}(u,1,i), \ldots , U_{l,k}(u,k-1,i) )$ allows to go from $P_{l,k}(i_{1})$ to $P_{l,k}(i_{0})$.
For $k$ an even number, $i_0$ and $i_1$ have the same parity: $i_0$ is even iff $i_1$ is even.

\begin{figure}
\begin{center}
\includegraphics[width=0.99\textwidth]{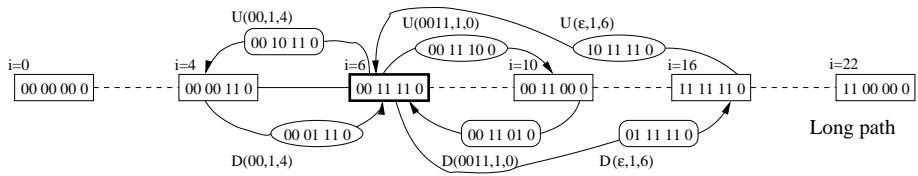} 
\caption{Extra paths linking the solution $P_{7,2}(6)$ of \mpath~of dimension $7$. 
Solutions in a rectangle are along the long path (\textit{i.e.} the efficient set).
Solutions in an ellipse are in the extra paths leading to solution $P_{7,2}(6)$ at the same position $(12.5, 22.5)$ in the objective space.
The solutions in a rounded rectangle are in extra paths beginning at the solution $P_{7,2}(6)$ translated by $(-0.5, -0.5)$ in the objective space to their destination solution. The length of extra paths is $1$.
Each solution is labelled by $D$ and $U$.
\label{fig:ex_path}}
\end{center}
\end{figure}

\begin{figure}
\begin{center}
\includegraphics[width=0.63\textwidth]{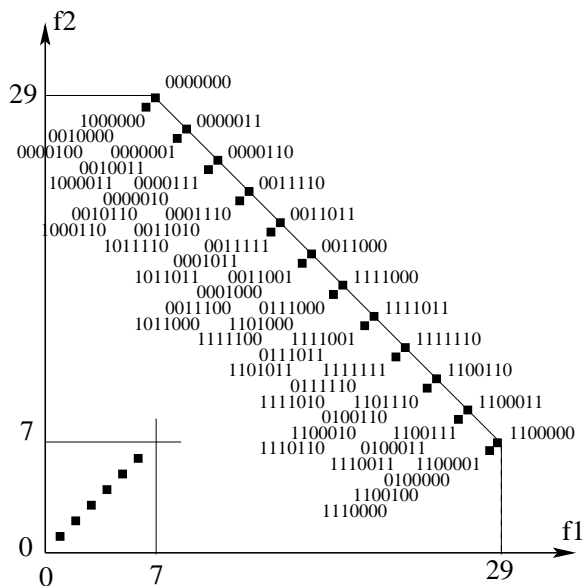} 
\caption{Objective space of the biobjective multiple $2$-path problem of dimension $l=7$. \label{fig:repr_m_path}}
\end{center}
\end{figure}

In \mpath, the efficient set corresponds to the set of solutions $P_{l,k}(i)$ in the long path where $i$ is an even number.
The efficient set is then fully disconnected with respect to Hamming distance $1$.
Solutions $P_{l,k}(2n+1)$ which are out of the efficient set are translated by a vector $(-0.5, -0.5)$ `under' the solutions $P_{l,k}(2n+2)$, so that they become dominated.
As a consequence, a solution $P_{l,k}(2n+1)$ leads to, but is dominated by, the efficient solution $P_{l,k}(2n+2)$.
However, $P_{l,k}(2n+1)$ and $P_{l,k}(2n)$ are mutually non-dominated.
In the same way, 
the extra paths to go from $P_{l,k}(i_0)$ to $P_{l,k}(i_1)$ are put on the first diagonal of the square enclosed by $(x_{i_1} - 1, y_{i_1} - 1)$ and $(x_{i_1}, y_{i_1})$.
More formally, the fitness function of the \mpath~can be defined as follows.
For all $x \in \{0,1\}^l$:
$$
f(x) =
\left\lbrace
\begin{array}{l l}
h_{l,k}(i) & \text{if } x \in P_{l,k} \text{ and } x=P_{l,k}(i) \text{ and } i \text{ even} \\
h_{l,k}(i+1) - (0.5 , 0.5) & \text{if } x \in P_{l,k} \text{ and } x=P_{l,k}(i) \text{ and } i \text{ odd} \\
h_{l,k}(i_1) - (\frac{k-j}{k}, \frac{k-j}{k}) & \text{if } x \in D_{l,k} \text{ and } x \not\in P_{l,k} \text{ and } \\
                                                     & x=D_{l,k}(u,j,i) \text{ with } P_{l,k}(i_1) = u 1^k P_{l,k}(i) \\
h_{l,k}(i_0) - (\frac{k-j}{k}, \frac{k-j}{k}) & \text{if } x \in U_{l,k} \text{ and } \\
                                                     & x=U_{l,k}(u,j,i) \text{ with } P_{l,k}(i_0) = u 0^k P_{l,k}(i) \\
(|x|_{0}, |x|_{0})  & \text{otherwise}\\
\end{array}
\right.
$$
Fig. \ref{fig:ex_path} illustrates the extra paths starting from one solution.
Fig. \ref{fig:repr_m_path} shows the objective space of a \mpath~instance.
For $j < k - 1$, solution $D_{l,k}(u,j,i)$ is a neighbor of solution $D_{l,k}(u,j+1,i)$ and is dominated by it.
As well, solution $D(u,k-1,i)$ is a neighbor of the efficient solution $P_{l,k}(i_1)$ and is dominated by it.
However, all $D_{l,k}(u,j,i)$ and $P_{l,k}(i_0)$ are mutually non-dominated.
The extra paths $D$ (Down) lead to a further solution in the long path, 
and the extra paths $U$ (Up) are the backward paths of the extra paths~$D$.
With those extra paths, an algorithm based on one bit-flipping can reach an efficient solution easily,
just by following the sequence defined by the set of mutually non-dominated solutions found so~far.

\subsection{Experimental Analysis}

\begin{figure}
\begin{center}
\includegraphics[width=0.7\textwidth]{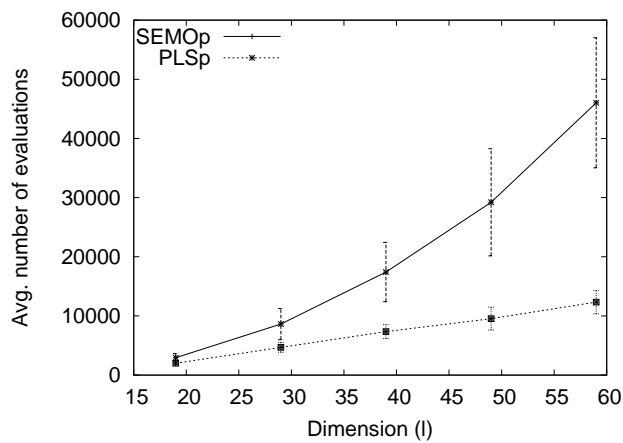}
\caption{Average value and standard deviation of the number of evaluations 
for PLS$_p$ and SEMO$_p$ on biobjective multiple $2$-path problems. 
\label{fig:multiplepath}}
\end{center}
\end{figure}

\begin{figure} 
\begin{center}
\includegraphics[width=0.7\textwidth]{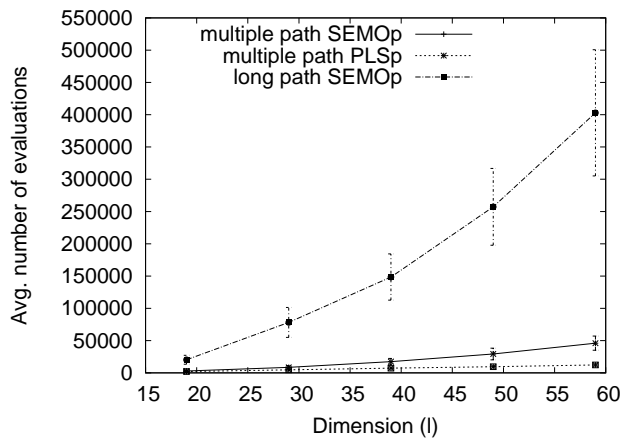} 
\caption{Average value and standard deviation of the number of evaluations 
for PLS$_p$ and SEMO$_p$ on biobjective multiple $2$-path problems  
compared to the SEMO$_p$ on biobjective long $2$-path.
\label{fig:multiplepath_compare}}
\end{center}
\end{figure}

The experimental study is conducted with the same approaches and parameters defined for the biobjective long path problem
on the previous section.
Fig. \ref{fig:multiplepath} shows the average value and the standard deviation of the number of evaluations for each algorithm.
Fig. \ref{fig:multiplepath_compare} allows to compare the number of evaluations with the previous problem.
Contrary to the results obtained for the long $2$-path problem, 
PLS$_p$ here clearly outperforms SEMO$_p$ which needs $3$ times more evaluations for dimension $l=49$. 
For PLS$_p$, the number of evaluations increases linearly with the dimension of the problem instance.
PLS$_p$ can find easily the same shortcuts than SEMO$_p$,
and the latter now loses computational resources to explore dominated solution and to evaluate the neighborhood of some solutions from the archive more than once. 
The curves on the right show that it is much easier to sample the efficient set of the multiple $2$-path than for the long $2$-path problem:
for dimension $49$, nearly $27$ times more evaluations are required between SEMO$_p$ for \lpath~and PLS$_p$ for \mpath.

This is the main results of this study.
The extra paths guide the search process to efficient solutions distributed all over the Pareto front.
The extra solutions are not in the efficient set and do not appear on the graph of efficient solutions,
but they are the keys to explain the performances of local search approaches.
Indeed, efficient solutions can now be reached very quickly by following the extra paths,
this explains the good performances of the algorithms.
Features from the efficient set (connectedness, etc.) are independent of the solutions from the extra paths.
Hence, the features of the efficient set are not the only key issue to explain the success of local search for MoCO.

\section{Conclusions and Future Works}
\label{sec:discs}

In this paper, we proposed two new classes of biobjective combinatorial optimization problems, the long and the multiple path problems,
in order to demonstrate empirically that 
connectedness is not the only key issue that characterizes the difficulty of a multiobjective combinatorial optimization problem.
In other words, connectedness is not the `Holy Grail' of search space features when the efficient set is intractable,
and when the goal is to find a limited-size approximation.
Indeed, on the long path problems, 
where the efficient set is intractable and connected, 
our experiments show that the running time to approximate it is exponential for a Pareto-based local search (PLS),
and polynomial for a simple Pareto-based evolutionary algorithm (SEMO).
On the multiple path problems, where the efficient set is still intractable but disconnected, 
PLS now outperforms SEMO, which seems rather unexpected at first sight.
This suggests two new considerations to measure the difficulty of finding a good efficient set approximation:
\begin{itemize}
\item First, the structure of the graph of efficient solutions induced by the neighborhood relation defined by the algorithm should also be taken into account. In the long path problems, this graph is a huge line for PLS whereas it is highly connected for SEMO.
Extending the notion of cluster on the efficient graph as defined by Paquete and St\"utzle \cite{paquete2009},
we should study a graph where an edge between efficient solutions is defined as the probability to reach one solution from the other.
\item Second, the solutions outside the efficient set should also be considered. In the multiple path problems, some solutions outside of the efficient set are temporally non-dominated so that they are saved into the archive during the search process. They help to approximate the (disconnected) efficient set.
\end{itemize}
In some sense, the fitness landscape of biobjective multiple path problems is unimodal, with a number of short paths leading to good solutions.
On the contrary, the biobjective long path problem can be characterized by a unimodal landscape where the path to good solutions is intractable.

Clearly, following the work of Horoba and Neumann \cite{horoba2009},
the next step will consist in leading a rigorous runtime analysis of PLS and SEMO for both the multiple and the long path problems.
The actual bounded archiving method is probably too specific, and seems very difficult to study rigorously. 
Then, in order to do so, we certainly have to change this strategy with the concept of $\epsilon$-dominance, for instance.
It is also possible to extend the biobjective path problems proposed in this paper to a larger objective space dimension
(more than $2$ objective functions), 
or with a larger `disconnectedness' (delete more than one solution over two).
The next challenge will be to define a relevant definition of fitness landscape in order to better understand the difficulty of multiobjective combinatorial optimization problems.
Given that the goal is here to find a set of solutions, we believe that another way to do so would be to analyze a fitness landscape 
where the search space consists of sets of solutions.
A solution would then be a set of bit strings instead of a single bit string for the problems under study in this paper.
Therefore, we plan to formally define fitness landscapes for the recent proposal of \emph{set-based} multiobjective optimization~\cite{zitzler2010}.

\paragraph*{Acknowledgments.}
The authors are grateful to Dr. Dirk Thierens for useful suggestions on the relation between intractable efficient sets and long path problems. 
They would also like to thank Dr. Luis Paquete for fruitful discussion on the subject of this work.

\bibliographystyle{splncs}

\end{document}